# Exploring approaches to computational representation and classification of user-generated meal logs

Guanlan Hu, PhD [*1,2], Adit Anand, MA [*1], Pooja M. Desai, MA, MPhil [1], Iñigo Urteaga, PhD [3], Lena Mamykina, PhD [1,2]

1 Department of Biomedical Informatics, Columbia University Irving Medical Center, New York 10032, USA
2 The Data Science Institute, Columbia University, New York 10032, USA
3 BCAM - Basque Center for Applied Mathematics & IKERBASQUE - Basque Foundation for Science. Bilbao, Spain

**Corresponding Author:** Lena Mamykina, PhD, Department of Biomedical Informatics, 622 West 168th Street, PH-20, New York, NY 10032, USA; om2196@cumc.columbia.edu

# ABSTRACT

**Objective:** This study examined the use of machine learning (ML) and domain-specific enrichment on patient-generated health data, in the form of free-text meal logs, to classify meals on alignment with different nutritional goals.

**Materials and Methods:** We used a dataset of over 3000 meal records collected by 114 individuals from a diverse, low-income community in a major US city using a mobile app. Registered dietitians (RDs) provided expert judgement for meal-goal alignment, used as "gold-standard" for evaluation. Using text embeddings (TF-IDF and BERT) and domain-specific enrichment information (ontologies, ingredient parsers, and macronutrient contents) as inputs, we evaluated the performance of logistic regression and multilayer perceptron classifiers using accuracy, precision, recall, and F1 score against the gold standard and individual's self-assessment.

**Results:** On average, individuals who logged meals achieved 0.576 accuracy of meal-goal alignment self-assessments. Even without enrichment, ML outperformed individual's self-assessments, with accuracies within 0.726-0.841 for different goals. The best-performing combination of ML classifier with enrichment achieved even higher accuracies (0.814-0.902). In general, ML classifiers with enrichment of Parsed Ingredients, Food Entities, and Macronutrients information performed well across multiple nutritional goals, but there was variability in the impact of enrichment and classification algorithm on accuracy of classification for different nutritional goals.

**Conclusion:** ML can utilize unstructured free-text meal logs and reliably classify whether meals align with specific nutritional goals, exceeding individuals' self-assessments, especially when incorporating nutrition domain knowledge. Our findings highlight the potential of ML analysis of patient-generated health data to support patient-centered nutrition guidance in precision healthcare.

*INTRODUCTION*

Patient-generated health data (PGHD), captured by patients through devices such as smartphones, wearables, and cameras, offers unique, context-rich insights that complement traditional clinical data. PGHD serves as a critical source of information for evidence-based healthcare, with promising potential to benefit high-quality and patient-centered precision care [1]. The use of PGHD can support clinical outcome assessments, enhance diagnosis and management of chronic diseases, reduce patient burden, foster innovation in medical tools and devices development, and promote health equity [2-6].

Despite its potential, effectively utilizing PGHD for clinical purposes remains challenging [5]; this is particularly the case for the emerging field of precision nutrition. In this domain, PGHD includes diverse data types, including, for example, food diaries, meal logs, physical activity self-tracking, sleep patterns, blood glucose measurements, as well as dietary preferences and restrictions, many of which are recorded as unstructured free-text entries. Such free-text data is often sparse, inconsistent, and can be highly variable in language, spelling, and cultural references, complicating standardization and integration with existing clinical data. Mobile apps for diet tracking, which often rely on image capture and brief unstructured text, suffer from similar problems and can produce idiosyncratic, error-prone logs lacking standardization [7]. Common approaches to addressing these challenges include structured digital logging through food databases or standard questionnaires, which enable collection of data more amenable to computational analysis. However, these approaches are time-consuming, error-prone, and burdensome for patients. These limitations underscore the need for innovative solutions to effectively analyze free-text PGHD.

Advancements in machine learning (ML) and natural language processing (NLP) offer a promising solution to address these challenges. For example, these methods can help to automatically and cost-effectively pre-process unstructured text and integrate heterogeneous data sources. Furthermore, enriching patient-entered records with domain-specific knowledge in the form of ontologies and nutritional databases can further facilitate computational analysis of nutritional records. This enhanced analysis enables personalized dietary recommendation [8], timely glucose control [9], and effective weight management. At a broader level, it supports healthy eating habits, optimizes healthcare resource allocation, reduces healthcare costs, and informs evidence-based public health interventions.

However, while current research demonstrates the successful application of NLP and ML in nutrition data analysis, their implementation for PGHD data, especially unstructured text, in precision nutrition remains under-explored. Previous studies have primarily focused on structured or semi-structured sources, such as food images, standardized nutrition labels, and recipes; linking food descriptions with nutrient fact tables to enable inferences from data to predict categorization, nutrition quality, and clinical outcomes such as blood glucose. For instance, NLP and ML techniques have effectively automated tasks such as food and recipe classification, ultra-processed food classification, and nutrition quality prediction [10-15]. Additionally, ML and generative AI models have also been applied to analyze dietary patterns and build personalized nutrition recommendation systems based on structured databases [16-21]. However, no study to date has utilized ML or NLP to address the specific challenges posed by unstructured PGHD data, particularly unstructured, patient self-reported dietary behavior entries, such as free-text meal records. While the potential of ML-NLP-driven analysis of unstructured data extends beyond nutrition, including applications to mental health (e.g., patient-reported mood and stress) and disease phenotyping, detection and prevention [22 23], few studies have systematically explored methods tailored explicitly for PGHD.

This study addresses this gap by evaluating the extent to which ML models, combined with domain-specific enrichment techniques and NLP methods for processing text, can accurately classify meals on their alignment with different nutritional goals using unstructured meal log text. We chose this classification task because it is an essential task in many nutritional interventions but can be challenging for non-experts. Based on a dataset of over 3000 meal records collected with a mobile app by 84 individuals in a diverse, low-income U.S. community, we compare model predictions with gold-standard assessments from registered dietitians. Our specific research questions include the following:

1. Q1: Can ML accurately classify meals logged in free-text format as meeting or not meeting specified nutritional goals?

2. Q2: How does ML classification accuracy compare to accuracy of such assessment by individuals who recorded the meals?
3. Q3: Can classification accuracy be improved with addition of different text enrichment techniques and different types of embeddings?
4. Q4: Are there particular types of enrichment techniques/embeddings that lead to improved classification accuracy across nutritional goals and classification methods?

**MATERIALS AND METHODS**

*Data Source*

Our dataset contains meal records collected from two previous studies by the research team. The first study was conducted in 2018-2019 and included a pilot assessment of a chatbot for personalized coaching in type 2 diabetes (T2 Coach) [24]. The second study was conducted in 2017-2019 and examined users' engagement with a novel mHealth app for diabetes self-management, Platano [25]. In each study, participants logged meals using a standard interface that captured: (1) meal type (breakfast, lunch, dinner, or snack), (2) a free-text meal title, (3) a free-text description of the meal (as detailed as participants wished), and (4) a photograph of the meal captured with their smartphone. In contrast to past research that already examined approaches to meal analysis using images [26], in this study, we specifically focus on using free-text user-entered descriptions.

Both studies, Platano and T2 Coach, included features for setting nutritional goals for healthy eating; the goals were developed by a team of Registered Dietitians (RDs) and consistent with recommendations by the United States Department of Agriculture (USDA) MyPlate [27]. Examples of such goals include "Choose lean proteins" and "Drink water with your meals"; see a description of the full list of nutritional goals in **Supplementary Table 1**.

Participants of the studies were asked to choose a nutritional goal from the offered list and then assess each newly logged meal on its alignment with their selected nutritional goal. In addition to these self-assessments, a team of six registered dietitians used a unified protocol to evaluate each logged meal and assign a binary label of whether the meal met the user's goal or not. The original dataset consisted of 5,284 unique meals recorded from 114 different individuals recruited from a low-income, ethnically diverse community in a major US city. Users could choose to submit meals in either English or Spanish. A subset of 3,169 of these meals recorded by 84 unique participants were logged in English and were used for model development and analysis.

*Nutritional Goals*

We selected four nutritional goals (out of 12 available in T2 Coach and Platano): "Drink water with your meals" (short name *drink water*, 431 labeled meals), "Choose lean proteins" (short name *lean protein*, 498 labeled meals), "Make half of your meal fruits and vegetables" (short name *half fruits and vegetable*, 369 labeled meals), and "Make ¼ of your meal carbohydrates" (short name *one fourth carbohydrates*, 514 labeled meals). We chose these goals because they differ in semantic complexity and corresponding self-assessment effort. Specifically, *drink water* simply requires the presence of water in the meal (or the word "water" in the meal description). *Lean protein*, however, is more complex, because it requires some inference of which foods are considered lean proteins, and the words "lean protein" are unlikely to be present in the meal description. Both goals are qualitative in nature as they do not consider quantities of particular foods. In contrast, the other two goals are quantitative; furthermore, both require additional inference to identify which items included in meal descriptions can suggest the presence of "fruits and vegetables" and "carbohydrates".

*Data Preparation*

We converted all text to lowercase and applied a preprocessing pipeline to the meal records' free-text fields that corrected misspellings, filtered out stop words defined by Python's Natural Language Toolkit (NLTK) [28], and removed irrelevant characters (i.e., punctuation, numbers, abbreviations, etc.). Using this cleaned data, we concatenated the meal title and meal description into a single text input. This raw text (referred to as "No Enrichment") was used as our baseline.

*Enrichment Techniques*

*Nutritionix.* We leveraged Nutritionix, an NLP toolkit (Syndigo, Inc) [29], for computing nutritional content from free-text input. Nutritionix maintains a database of over 1,200,00 food items, which includes entries for restaurant and packaged foods, along with their respective nutritional information. Nutritionix publishes an API that allows users to extract food entities from inputted text data, learn the ingredient composition of a food entity, and query an entity's nutritional information. Example outputs from the Nutritionix API can be found in **Supplementary Table 2**.

*FoodOn.* We utilized FoodOn, a food ontology developed by agencies across Canada, United Kingdom, and United States to standardize food concept usage in domains such as nutrition and food composition [30]. FoodOn is structured using concept relationships ranging from food categories and products, to packaging and preservation processes. Since FoodOn is built upon vocabularies derived from the intersection of science and food regulation domains, it is a useful public ontology to tackle challenges in precision nutrition problem spaces. We illustrate in **Figure 1** the ontological representation of the FoodOn concept "corn flakes."

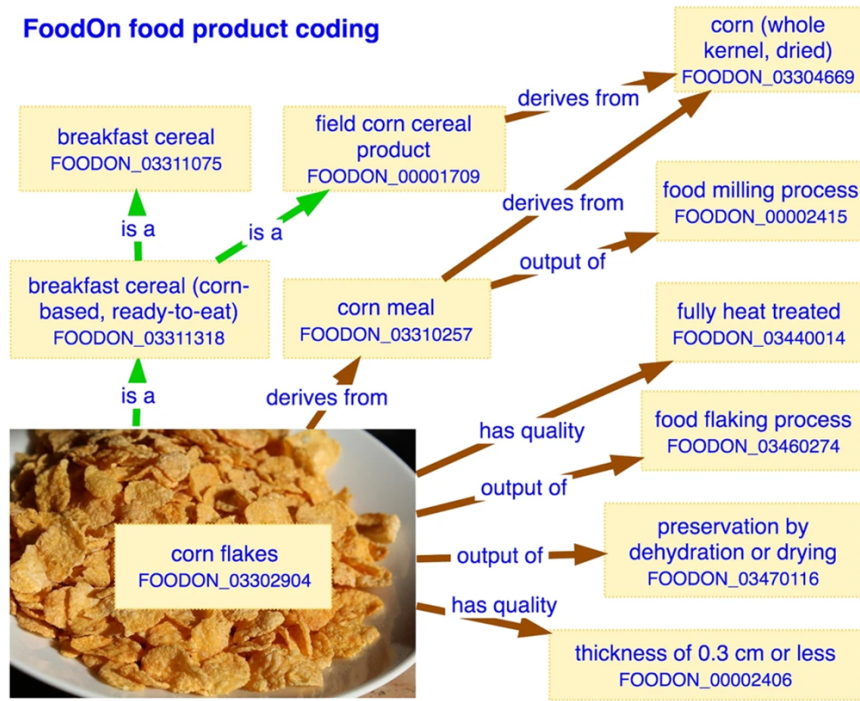

**Figure 1. A diagram highlighting the ontological relationships and hierarchy of the FoodOn concept for "corn flakes."**

*Engineered Features*

We use the Nutritionix API to determine the food entities (food ingredients) contained in the meal record title and description. We also retrieve the component ingredients contained within each food entity (parsed ingredients) using the Nutritionix API (i.e., the entity "salad" contains ingredients such as "spinach" and "iceberg lettuce"). The Nutritionix API also provides the macronutrient information (calories, carbs, dietary fiber, fat, and protein) contained in each meal record (referred to as macronutrients text). Using the macronutrients text data, we engineer a feature that captures the normalized, numeric sums of a meal record's macronutrient content (macronutrients numeric). For each food entity identified by the Nutritionix API, we use FoodOn ontology to retrieve the corresponding ontological hierarchy and concepts, and concatenate these concepts into a text representation (referred to as FoodOn).

Based on this plethora of information, we created various combinations of enrichment techniques. **Supplementary Table 3** is an example to demonstrate how each technique enriches a sample free-text meal record (i.e., "fried whiting sandwich wheat bread bottle water").

*Embeddings and Classification Algorithms*

For each enrichment technique, we concatenate the meal title and meal description with enriched text and represent them computationally via word embeddings using TF-IDF and BERT. Term Frequency Inverse Document Frequency (TF-IDF) is a statistical method that reflects a word's importance within a document relative to a corpus [31]. It is used to transform the text into a high-dimensional sparse feature vector based on weighted term frequencies. Bidirectional Encoder Representations from Transformers (BERT) is a pre-trained language model that captures contextual relationships between words by considering both left and right context. It is pre-trained to capture semantic meanings of words more accurately [32]. BERT-base were applied to convert meal records text into numerical representations. We considered both TF-IDF and BERT to compare frequency-based lexical features with semantic embeddings, allowing us to evaluate whether contextualized representations provide improved performance for classifying nutritional goal alignment.

Next, we trained two supervised binary classifiers on the text embeddings to predict whether each meal aligned with a given nutritional goal. We selected two representative approaches: a simple, linear decision rule-based classifier (Logistic Regression) and a more advanced, neural network-based nonlinear classifier (Multilayer Perceptron). The Logistic Regression (LR) is a supervised machine learning algorithm for binary classification, which predicts the probability of an outcome being equal to 1 or 0 [33]. The Multilayer Perceptron (MLP) is a feed-forward neural network with one or more hidden layers, which uses ReLU activations, and is trained via stochastic gradient descent for optimal classification performance [34].

*Evaluation Strategy*

To evaluate the performance of the classification algorithms, we used accuracy (ratio of correctly predicted observations to the number of total observations), precision (ratio of true positive predictions to the total predicted positives), recall (ratio of true positive predictions to all actual positives), and F1 score (harmonic mean of the precision and recall). Data were split into 80% (training) and 20% (testing) sets. All the analyses were conducted using Python 3.9.

**RESULTS**

*Individuals and RDs meal-to-goal alignment assessment across all nutritional goals*

To contextualize model performance, we first examined how well individuals' self-assessments aligned with RDs expert evaluations, measured by the percentage of decisions that matched across all nutritional goals in our dataset, i.e., not only the four goals selected for subsequent analysis. Overall, participants' assessments aligned with assessments by expert RDs on 57.6% of meals, with alignment rates ranging from 34.8% to 88.9% depending on the goal (**Figure 2**).

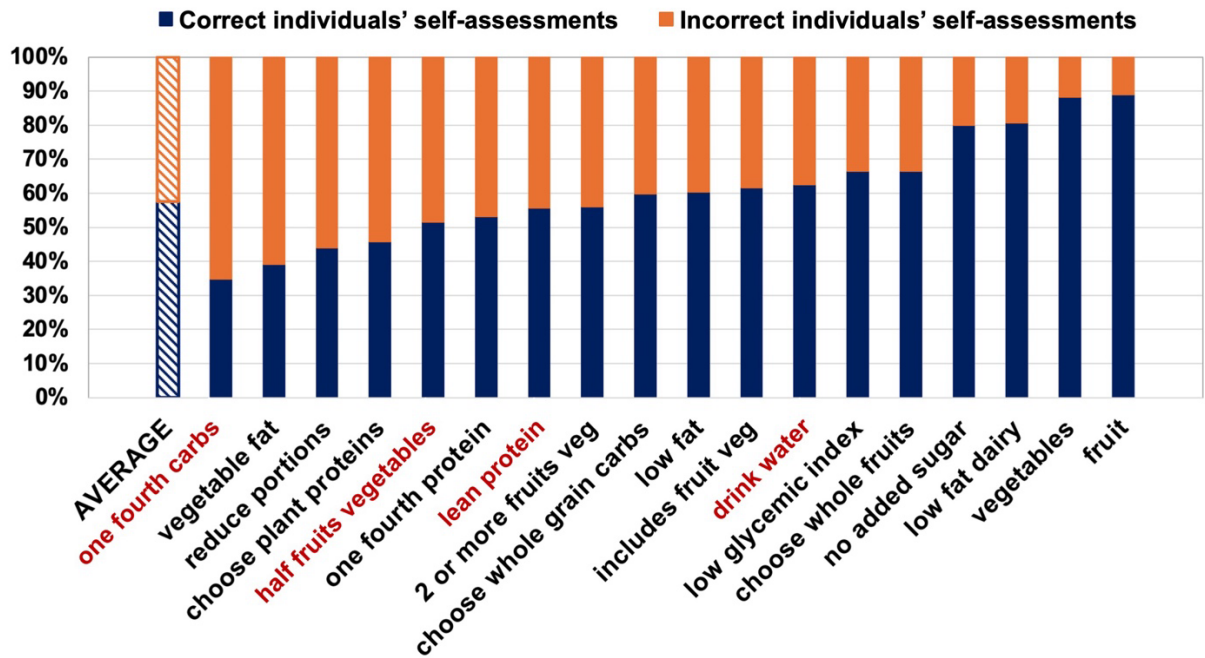

**Figure 2. Alignment between individuals' self-assessment and registered dietitians' (RDs) assessment across nutritional goals (full descriptions for all nutritional goals is available in Supplementary Table 1).**

The lowest levels of alignment (below 50% accuracy) were observed in four of the nutritional goals that are more complex or ambiguous or require estimating portion sizes, such as *one forth carb* (34.8%), *vegetable fat* (39.1%), *reduced portions* (44.0%), and *choose plant proteins* (45.8%). Moderate agreement (accuracy 50%-70%) occurred for goals such as *half fruits and vegetables* (51.6%), *lean protein* (55.7%), *and drink water* (62.6%), indicating that while individuals generally understood these nutritional goals, some confusion or difficulty remained. At the higher end, about 80% or more of individuals' self-assessments aligned with RDs for nutritional goals such as *no added sugar* (79.9%), *low fat dairy* (80.7%), *vegetables* (88.2%), and *fruit* (88.9%).

Given the substantial variability and inconsistency across different goals in self-assessments, especially for goals requiring inference, we next evaluated whether machine learning models could achieve accurate and consistent classification performance.

*ML meal-to-goal classification, without enrichment, improves over individuals' self-assessments across nutritional goals*

To address Q1 and Q2—"Can ML accurately classify meals logged in free-text format as meeting or not meeting specified nutritional goals" and "How does ML classification accuracy compare to accuracy of such an assessment by individuals who recorded the meals"—we compared individuals' self-assessments performance (on the test set) to those of ML algorithms trained on RDs assessments.

Results show that even without enrichment, ML algorithms can accurately classify goal alignment. Specifically, our models reached 0.765-0.827 accuracy and 0.750-0.799 F1 score for *drink water* goal, 0.801-0.841 accuracy and 0.539-0.730 F1 score for *lean proteins* goal, 0.729-0.759 accuracy and 0.342-0.603 F1 score for *half fruits and vegetables* goal, 0.726-0.784 accuracy and 0.000-0.328 F1 score for *one forth carbohydrates* goal (**Figure 3** and **Supplementary Table 4**).

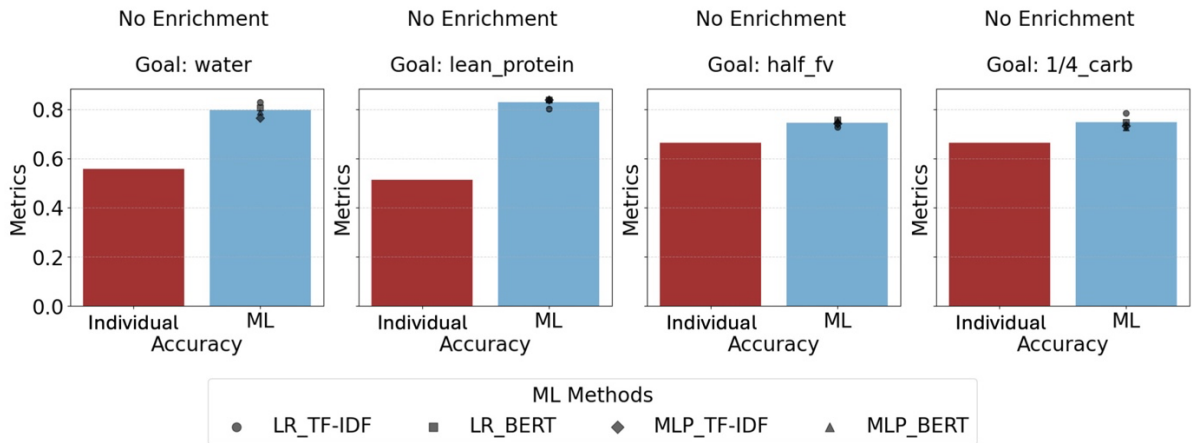

**Figure 3. ML meal-to-goal classification algorithms improve individuals' self-assessments across selected nutritional goals.**

Notably, all four combinations of ML methods and embedding techniques (logistic regression with TF-IDF, logistic regression with BERT embeddings, multilayer perceptron with TF-IDF, and multilayer perceptron with BERT embeddings) consistently achieved robust and accurate performance, with model-specific performance markers shown within each blue bar in **Figure 3**.

Compared to individuals' self-assessments, ML algorithms achieved higher decision accuracy across the four goals, even without any enrichment of individuals' recorded meal descriptions (Figure 3). For each goal, ML predictions (blue bars) consistently outperformed individuals' self-assessments (red bars), as measured by accuracy of ML classifications compared to gold standard (RDs assessments). For the drink water and lean proteins goals, ML methods showed substantial improvements in classification accuracy compared to individuals' self-assessments, rising from approximately 0.50 to around 0.80 (0.726 to 0.841). For the half fruits and vegetables and one forth carbohydrates goals, ML methods still outperformed self-assessments, though the gains were more modest, with accuracy increasing by less than 0.10.

*ML with enrichment achieves higher performance, with variability across different goals and enrichment methods*

To address Q3, we evaluated the potential of the different input enrichment techniques to further improve the performance of ML algorithms in assisting individuals' assessments on their nutritional goals. This analysis showed 3 key findings (**Figure 4 and Supplementary Table 4**):

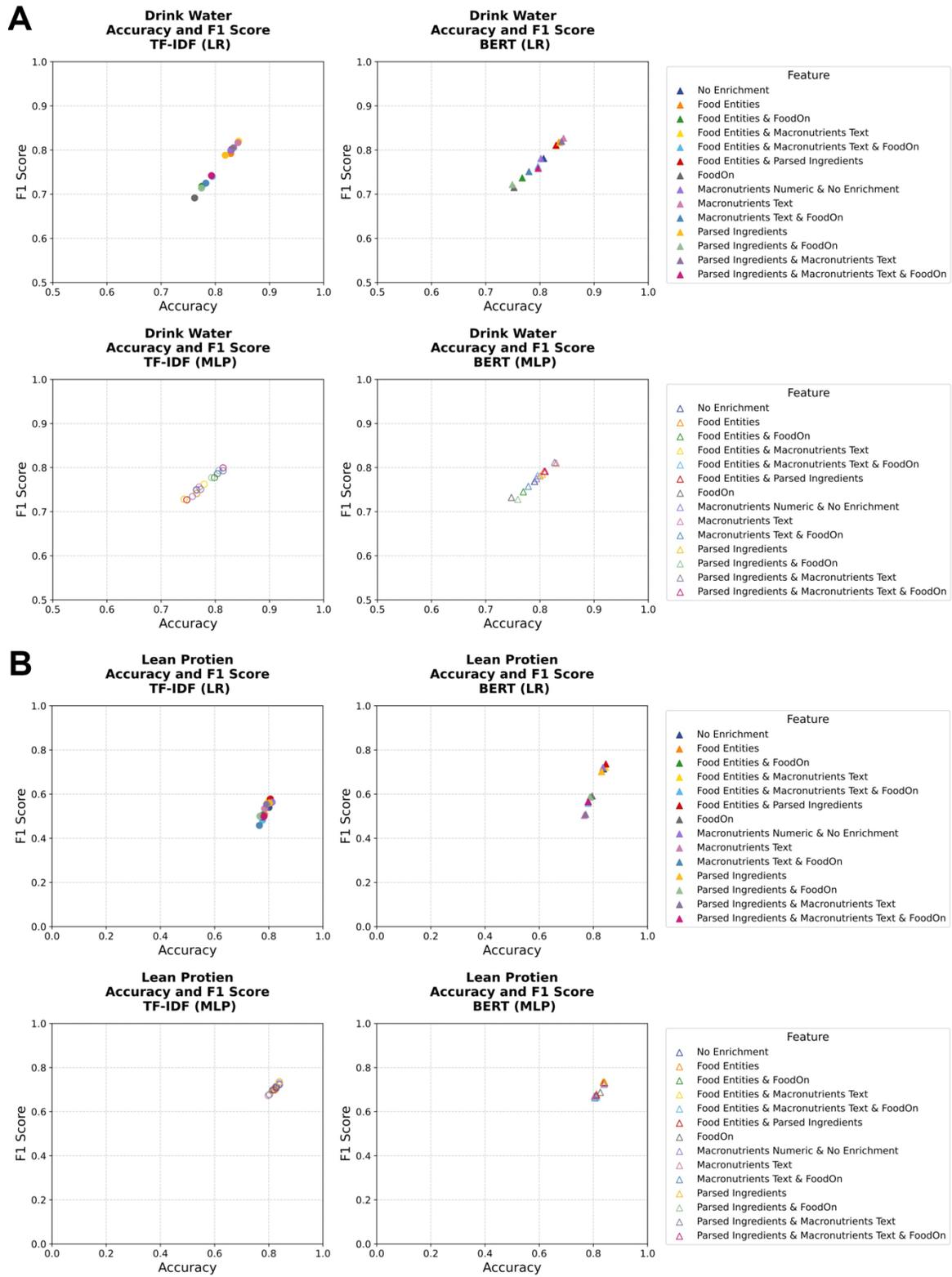

**Figure 4. ML predictions for (A) the *drink water* goal, (B) the *lean protein* goal.**

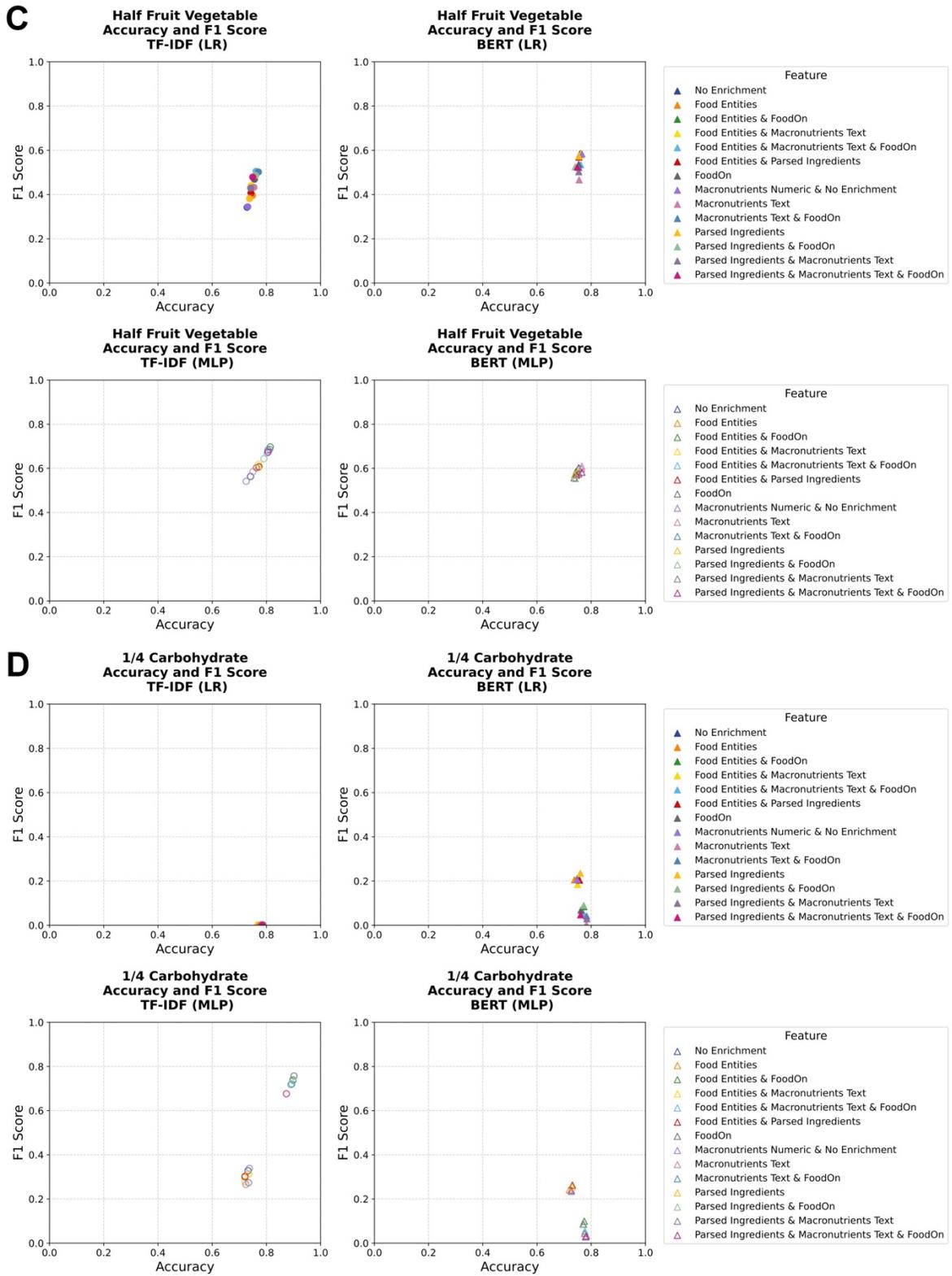

**Figure 5. ML predictions for (C) the *half fruits and vegetable* goal, and (D) the *one fourth carbohydrates* goal.**

1. ML with enrichment achieves higher performance

Overall, introduction of enrichment techniques improved the performance of ML methods, across all nutritional goals. Results revealed that combining MLP TF-IDF with FoodOn achieved the highest accuracy of 0.90 (F1 score 0.76), while the LR BERT with Macronutrients Text achieved the highest F1 score of 0.83 (accuracy 0.84).

2. Different enrichment techniques show different performance across different goals

For the *drink water* goal, LR BERT with *Macronutrients Text* achieved the highest prediction accuracy of 0.8437. *Food Entities, Food Entities & Macronutrients Text, Food Entities & Parsed Ingredients, Macronutrients Text, Parsed Ingredients,* and *Parsed Ingredients & Macronutrients Text* all performed well, with most ML methods exceeding 0.80 in prediction accuracy.

When predicting the *lean protein* goal, LR BERT with *Food Entities & Parsed Ingredients* reached the highest accuracy of 0.846. *Food Entities & Parsed Ingredients, Macronutrients Numeric & No Enrichment, Parsed Ingredients,* and *No Enrichment* demonstrated strong overall prediction performance (>0.80 for all ML methods), followed by *Food Entities, Food Entities & Macronutrients Text* (>0.80 in 3 out of 4 ML methods), compared to other enrichment techniques.

For the nutritional goal of *half fruit and vegetables*, MLP TF-IDF with *Food Entities & FoodOn* achieved the highest accuracy of 0.8143. Only *Macronutrients Text & FoodOn* (0.8122), *Parsed Ingredients & Macronutrients Text & FoodOn* (0.8062), *FoodOn* (0.8059), and *Food Entities & Macronutrients Text & FoodOn* (0.803), all with MLP TF-IDF, exceeded 0.80 in accuracy. Other enrichment techniques across different ML methods performed relatively poorly.

Regarding the nutritional goal of *one fourth carbohydrates*, MLP TF-IDF with FoodOn attained the highest accuracy of 0.9017. Other than that, MLP TF-IDF with only *Food Entities & Macronutrients Text & FoodOn, Macronutrients Text & FoodOn, Parsed Ingredients & FoodOn, and Parsed Ingredients & Macronutrients Text & FoodOn* perform relatively better (>0.80) than all other enrichment techniques. Of note, this nutritional goal exhibited lowest precision, recall, and F1 score compared with other nutritional goals, and MLP demonstrated better F1 score compared to LR classifiers (see **Supplementary Table 4**).

***Some enrichment techniques consistently deliver higher accuracy across multiple ML algorithms.***

Finally, to address RQ4, we compared performance across different enrichment techniques and classification algorithms. Despite the differences described above, some enrichment techniques consistently delivered higher accuracy across multiple ML algorithms. Specifically, *Parsed Ingredients* and *Food Entities & Parsed Ingredients* showed strong overall performance, with a high chance of achieving above 0.80 accuracy (7 out of 16 combinations of ML methods and nutritional goals).

Several enrichment approaches, including *No Enrichment, Food Entities, Food Entities & Macronutrients Text, Macronutrients Numeric & No Enrichment,* and *Parsed Ingredients & Macronutrients Text & FoodOn*, each had 6 out of 16 methods that surpassed the 0.80 accuracy threshold. However, *Parsed Ingredients & FoodOn* had only 3 out of 16 methods that exceeded the 0.80 accuracy threshold.

On the other hand, *Food Entities & Macronutrients Text* enrichment techniques exhibited the best universal prediction performance among different ML algorithms, with 4 out of 16 combinations of ML methods and nutritional goals ranked as top 3 for accuracy. *Macronutrients Text* is the second optimal option, with 3 out of 16 combinations achieving the top 3 ranking methods in terms of accuracy.

## *DISCUSSION*

In this study, we demonstrated that machine learning (ML) methods, when combined with domain-specific enrichment techniques, can classify whether meals meet specific nutritional goals using patient-generated health data (PGHD) such as free-text meal records. These models significantly outperformed individuals' self-assessments, achieving an average accuracy score of 0.8 compared to 0.5 for participants (see **Figure 3**).

Overall, our findings revealed three key trends. First, ML models, even without enrichment, outperform individuals' self-assessments across different nutritional goals. Second, using different enrichment techniques resulted in considerable improvements in classification accuracy, with improvement gains reaching 0.719-0.902 and an average accuracy of 0.785. These improvements were particularly substantial for complex nutritional goals, such as the *one fourth carbohydrates* goal. Third, different enrichment techniques resulted in different accuracy gains across different goals; however, several enrichment techniques showed consistently good performance. For example, *Food Entities & Macronutrients Text*-based enrichment showed the strongest universal prediction performance among enrichment combinations across multiple nutritional goals, while *Food Entities & Parsed Ingredients* also exhibited consistently good performance. Overall, these findings show that unstructured PGHD (e.g., free-form meal logs) with ML analysis and enrichment can be effectively used to make important inferences regarding nutritional choices.

These findings extend previous investigations concerning the use of PGHD in health and wellness. While the challenges of PGHD including noise, inconsistency, and lack of standardization are well established in the literature, most prior work has focused on integrating structured or semi-structured PGHD with electronic health records (EHR) to support retrospective phenotyping or population-level analyses. For example, prior work has demonstrated the potential of machine learning and unsupervised methods to derive phenotypes from self-tracked symptom data in chronic conditions such as endometriosis, using longitudinal free-text entries collected through mobile health tools [35]. Other studies have explored the extraction of social and behavioral determinants of health from free-text EHR narratives using NLP techniques [36]. Similarly, free-text data from social media has been classified to infer mental health status, such as depression, based on language patterns [37 38]. While these efforts center on diagnostic inference or structured phenotyping, our approach diverges in both objective and granularity: we transform naturalistic, meal-level free-text logs into interpretable signals that can enable data-driven self-management interventions. In doing so, we illustrate how task-specific representation learning, when grounded in domain knowledge, can help bridge the gap between noisy, self-reported data and actionable support for everyday health behaviors.

Real-time assessment of meal-goal alignment bridges a critical gap in precision nutrition by enabling timely, patient-facing feedback and behavioral guidance based on free-text meal logs. This study expands the scope of computational PGHD research beyond retrospective clinical characterization, offering a foundation for scalable, personalized dietary interventions. Accurate assessment of alignment between an individual's meals and their nutritional goals is fundamental to their ability to follow healthful diet [39]. Yet previous research showed that this critical step presents insurmountable barrier to many individuals struggling with chronic diseases. For example, past research showed that individuals often underestimate portion sizes in their meals and overestimate consumption of fruits and vegetables [40]. This is particularly the case for individuals with low health and nutritional literacy [41]. Our approach on meal-goal alignment can provide these individuals with useful feedback and improve their chances of reaching their goals, supporting more effective dietary self-management in individuals with type 2 diabetes. This capability is especially valuable for underserved populations, where traditional dietetic support may be limited, and digital tools offer an accessible alternative. Furthermore, this approach can pave the foundation for other types of nutritional interventions. For example, adding interpretability mechanisms and analysis to the classifier used in this study can help to identify meal features that contribute to misalignment with goals and suggest ways to change meals to improve meal-goal alignment. Incorporating features engineered using nutritional databases and ontologies can further enhance this ability.

Our study revealed substantial variability in classification performance across nutritional goals, highlighting both the complexity of certain tasks and the importance of task-specific design. As expected, different goals posed varying levels of difficulty in assessing meal-goal alignment. To reflect this diversity, we intentionally selected goals that differed in type and complexity. The results confirmed our hypothesis: alignment between participants' self-assessments and registered dietitian (RD) evaluations varied widely across nutritional goals, ranging from 34.8% for the *one fourth carbohydrates* goal to 88.9% for the *half fruits and vegetable* goal. Similarly, machine learning model performance and the impact of enrichment techniques differed by goal. For example, the *Food Entities & Macronutrients Text* enrichment yielded strong accuracy for *drink water* (0.8437 with BERT) and *lean protein* (0.8446 with BERT) but was less effective for "half fruits and vegetables" (0.7435 with TF-IDF) and particularly inconsistent for "one-fourth carbohydrates," where many enrichment methods failed to improve over the baseline without enrichment. These results highlight that no single enrichment approach performs optimally across all goal types, suggesting that tailoring models to the complexity and structure of specific dietary goals may be necessary to maximize classification accuracy.

This variability raises important questions regarding the trade-offs between cross-goal generalizability and goal-specific optimization for solution design. Generalizable solutions are easier to scale, as they can be trained on larger, cross-goal datasets and are more robust to class imbalance across goals. However, our study showed that these solutions will likely lead to lower accuracy in assessment of meal-goal alignment. In contrast, goal-specific solutions, though less scalable and requiring additional re-training for each newly added goal, enable tailored feature engineering, model tuning, and enrichment strategies that can substantially boost performance for individual goals. A key direction for future work lies in hybrid modeling strategies: can we develop modular or multitask architectures that retain the scalability of general models while incorporating goal-specific nuances? Addressing this question will be critical for building flexible, accurate, and interpretable nutrition-feedback systems at scale.

Despite these promising results, there are several limitations. First, the quality of PGHD is highly dependent on individuals' ability and willingness to accurately log dietary details, leaving records susceptible to errors, omissions, and inconsistencies. Secondly, our model was trained exclusively on English free-text entries, so its adaptability to multilingual or culturally diverse populations requires further exploration. Thirdly, we used a binary classification framework for determining whether a meal meets a goal or not, although multi-class classification may capture more complete nutritional goals. Future work could address these limitations by using more complex classifiers or multi-class classification methods, as well as using large language models (LLMs) to incorporate experts' explanations.

## *CONCLUSIONS*

This study highlights the potential of applying machine learning methods, particularly when enriched by domain-specific knowledge (i.e., ontologies and nutritional composition), in reliably transforming free-text meal logs into actionable feedback on whether individuals' meals have met their nutritional goals. Compared to individuals' self-assessment, ML models achieved significantly higher accuracy and consistency. While no single enrichment technique was optimal across all nutritional goal types, the use of *Parsed Ingredients*, *Food Entities*, and *Macronutrients Text* with BERT or TF-IDF embeddings proved beneficial for complex nutritional goals, which help individuals make more informed dietary choices. These findings underscore the potential of AI-enhanced PGHD analysis to advance evidence-based patient-centered nutrition guidance, particularly important for underserved populations with diabetes who might benefit most from accessible and tailored nutrition support.

**FUNDING**

This work was funded in part by an award from the National Science Foundation (Award number SCH:2306690) National Institute of Diabetes and Digestive and Kidney Diseases (Award number R01DK113189), and the National Library of Medicine (Award number T15LM007079). Iñigo Urteaga acknowledges the support by MICIU/AEI/10.13039/501100011033 and the BERC 2022-2025 program funded by the Basque Government.

**AUTHOR CONTRIBUTIONS**

GH, AA, PD, IU, LM designed the study. AA and GH implemented the methods and conducted all the analysis. PD and IU contributed to conducting experiments and evaluation of the results. GH led the writing of the manuscript. IU and LM supervised the research. All authors edited the manuscript. All authors were involved in developing the ideas and drafting the paper.

**SUPPLEMENTARY MATERIAL**

Supplementary materials are available at Journal of the American Medical Informatics Association online.

**ACKNOWLEDGEMENTS**

# CONFLICT OF INTEREST STATEMENT

None declared.

# DATA AVAILABILITY

The data underlying this study cannot be shared publicly due to the patient's privacy.